\documentclass[12pt,fleqn]{article}
\usepackage[small,bf]{caption}
\usepackage{graphicx,color}
\usepackage{amsfonts}
\usepackage{amssymb}
\usepackage{amsmath}
\usepackage{indentfirst}
\usepackage{natbib}
\usepackage{fancyhdr}
\usepackage[utf8]{inputenc}                              	  
\usepackage{bbm} 								
\usepackage{algorithm}						
\usepackage{algpseudocode}				
\usepackage{subfigure} 							
\usepackage{url}	
\usepackage[bookmarks,pdfborder={0 0 0}]{hyperref} 
\usepackage[top=2.0cm, bottom=3.3cm, left=2.5cm,right=2.5cm]{geometry} 
\usepackage{setspace}    					     
\definecolor{seagreen}{RGB}{46,139,87}
\usepackage{scalefnt}

\providecommand{\sin}{} \renewcommand{\sin}{\hspace{2pt}\mathrm{sen}}
 
\newcommand{\ie}{{\it i.e.}}

\newcommand{\eg}{{\it e.g.}}
\newcommand{\keywordsname}{Keywords}
\newenvironment{keywords}{%
\noindent
        {\em \bfseries \keywordsname: }%
         \rm }
      {\endquotation \vskip 12bp}

\begin{document}


\title{An Image-Based Fluid Surface Pattern Model} 

\author{\rm \small
\begin{tabular}{l}
\textbf{Mauro de Amorim}- mamorim@iprj.uerj.br\\
\textbf{Ricardo Fabbri}- rfabbri@iprj.uerj,br\\
\textbf{Lucia Maria dos Santos Pinto}- lmpinto@iprj.uerj.br\\
\textbf{Francisco Duarte Moura Neto}-  fmoura@iprj.uerj.br\\
Polytechnic Institute at the Rio de Janeiro State University\\
28630-050 - Nova Friburgo, RJ, Brazil\\
\end{tabular}
}

\maketitle

\lhead{\textcolor{blue}{\fontsize{9}{0pt}\usefont{T1}{phv}{m}{n}\it XVI Encontro de Modelagem Computacional \\ \vspace{1.0pt}
IV Encontro Ci\^encia e Tecnologia de Materiais \\ \vspace{1.0pt}
III Encontro Regional de Matem\'atica Aplicada e Computacional \\ \vspace{-4.0pt}
Universidade Estadual de Santa Cruz (UESC), Ilh\'eus/BA, Brasil. 23-25 out. 2013.}}
\chead{}
\rhead{}
\renewcommand{\headrulewidth}{0.0pt}
\fancyfoot{}

\begin{abstract}
This work aims at generating a model of the ocean surface and its motion
from one or more video cameras. The idea is to model wave patterns from video
as a first step towards a larger system of photogrammetric
monitoring of marine conditions for use in offshore oil drilling platforms.  The first
part of the proposed approach consists in reducing the dimensionality of sensor
data made up of the many pixels of each frame of the input video streams.
This enables finding a concise number of most relevant parameters to model the temporal dataset, yielding an
efficient data-driven model of the evolution of the observed surface.  The
second part proposes stochastic modeling to better capture the patterns
embedded in the data. One can then draw samples from the final model, which are
expected to simulate the behavior of previously observed flow, in order to determine
conditions that match new observations.  In this paper we focus
on proposing and discussing the overall approach and on comparing two different
techniques for dimensionality reduction in the first stage: principal component analysis
and diffusion maps. Work is underway on the second stage of
constructing better stochastic models of fluid surface motion as proposed here.
\end{abstract}


\section{Introduction}

The simulation of fluids is an imporant tool in computer graphics, \eg, for
generating realistic animations of water flow. Many fluid simulations perform
the evolution of liquid through time based on
the Navier-Stokes equation or simpler wave and frequency models, among others. This work is an
initial part of a larger effort to perform the inverse problem of generating graphics simulations, \ie,
to automatically extract 3D models of fluid surfaces starting from real-world
video data. The obtained model can then be used for simulating the motion of
ocean surface patterns as \emph{observed} in the real world. Moreover, this
simulation can be matched to video at new time instants in order to predict and
track the actual conditions of the \emph{observed} fluid.

In order to track the apparently complex dynamics of a large number of images of
the ocean surface, our proposed system reduces the dimensionality of previsouly observed data 
and automatically learns a concise data-driven model. This enables the inference
through synthesis of patterns that are intrisic to the observed phenomenon, while
drastically reducing the search space with little to no loss. The remaining
error in the low dimensional modeling is dealt with as part of a stochastic modeling
stage.

\begin{figure}[!httb]
\centering
\includegraphics[scale=0.2]{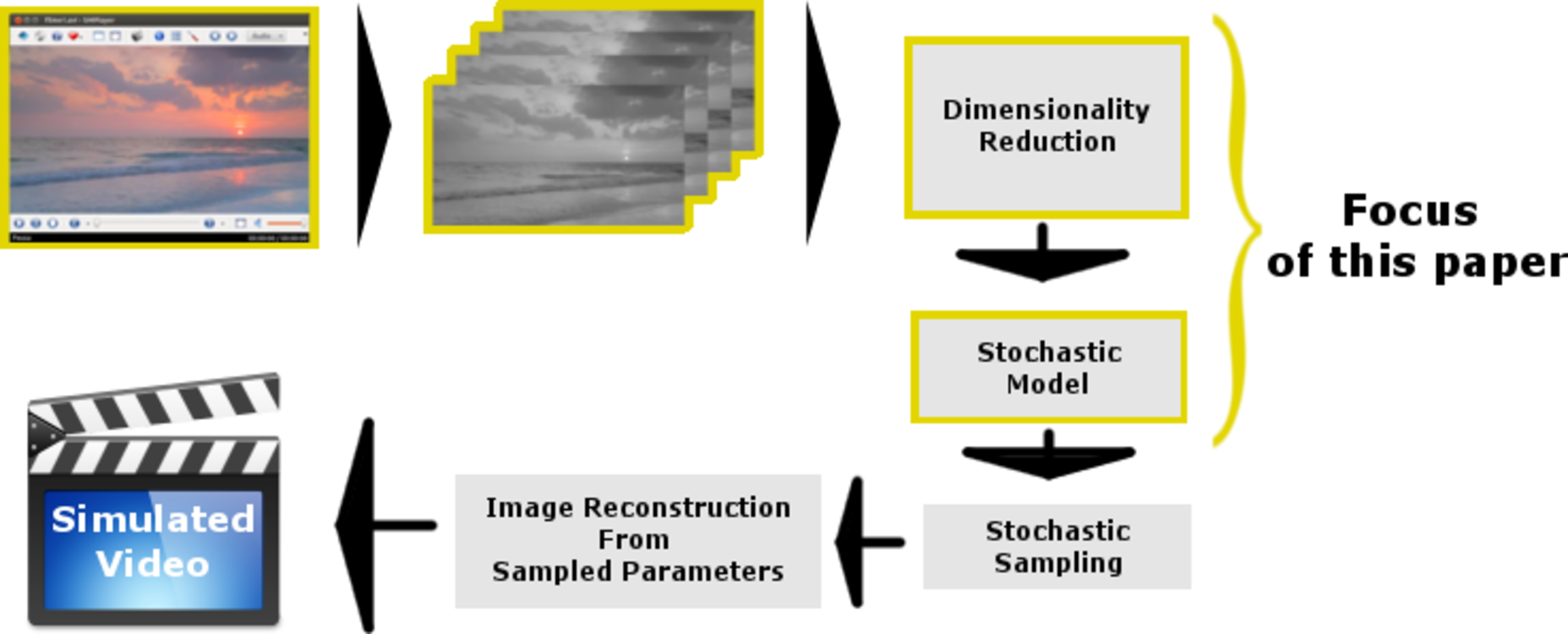}
\caption{Diagram of the proposed approach for the modeling of fluid surface
patterns aiming at applications of monitoring ocean conditions for
offshore oil drilling processes, among others.}
\label{fig:diagrama}
\end{figure}

While detailed techniques for the plausible inference involving such patterns is
ongoing work, we propose that the patterns in newly observed data are to be found by
stochastic modeling, using the Bayesian paradigm, a method which can be described
as ``analysis by synthesis''~\citep{bib:Mumford}.  Finally, a stochastic model
also enables generating plausible new images for a realistic image-based simulation.
Figure~\ref{fig:diagrama} shows a diagram representing the steps of the proposed
modeling approach dealt with in this manuscript.

Two dimensionality reduction techniques were assessed: \emph{(i)}
principal components analysis, which consists in performing orthogonal
projections of data along linear directions of greater variance; and
\emph{(ii)} diffusion maps, a recent non-linear technique which organizes data
into graphs of local similarity and extracts global structure through a diffusion
process~\citep{bib:Lafon}.

\section{Dimensionality Reduction Techniques}

Assume a video is made up of $n$ video frames represented by a column vector
$y_{i}$, $i=1,\dots,n$. Each video frame is an image with a total of $p$ pixels,
so that each observation or data point is a $p$-dimensional vector $y_i \in \mathbbm{R}^p$.
For instance, frames of a typical high-definition video $1920\times1080$ are each comprised of
more than 2 milion dimensions, an excessive number of parameters to capture the
observable modes of variation of fluid surface patterns that are relevant for
monitoring applications. We explore two
different techniques for automatically reducing this number of parameters as a step towards
generating a tractable yet meaningful model.

\subsection{Principal Component Analysis\label{sec:PCA}}

The principal component analysis (PCA) is a linear dimentionality reduction
technique. High-dimensional data, \ie, $p$-dimensional, gets reduced by PCA to a
global linear submanifold of reduced dimension $q \ll p$, determined by the directions
of greatest variability in the data. These directions are given by $q$
orthonormal vectors, the principal components, which are the eigenvectors
corresponding to the largest eigenvalues of the covariance matrix of the data.

Principal component analysis can be computed through the singular
value decomposition (SVD)~\citep{bib:Hastie,bib:Jolliffe,bib:Golub} as follows.
The $n$ observations $y_i \in \mathbbm{R}^p$ are mean-centered by subtracting
their average $\bar{y} = \dfrac{1}{n} \sum^{n}_{j=1} y_{j}$.
The new data vectors $x_{i} = y_{i}-\bar{y}$ form the columns of the $p\times n$
data matrix $X$ whose singluar value decomposition is given by
\begin{equation}\label{eq:SVD}
X = U \Sigma V^\top,
\end{equation}
with $U$ and $V$ being $p\times n$ and $n \times n$ orthonormal matrices, resp., 
and $\Sigma$ an $n \times n$ diagonal matrix, with diagonal elements $\sigma_1 \geq \dots \geq \sigma_n \geq 0$
known as singular values. The first $q$ columns of $U$ form the $p \times q$ matrix
$U_{q}$, the so-called first $q$ principal components,
and $\alpha_i = U_q\top x_i$ are the coordinates of any given observation $x_i$ in
the subspace of the principal components.

The computation of PCA using SVD is equivalent to computing the eigenvalues and
eigenvectors of the covariance matrix, which is given by
\begin{equation}
\dfrac{1}{n-1} X X^{\top}.
\end{equation}
It is possible to show this fact by taking the SVD of the matrix $X$ as shown
in~\eqref{eq:SVD}, and building the product $XX^{\top}$, giving 
\begin{equation}
X X^{\top} = (U \Sigma V^\top)(V	 \Sigma U^\top).
\end{equation}
Since $V$ is an orthogonal matrix, we have
\begin{equation}\label{eq:rel_SVD}
X X^{\top} = U \Sigma^2 U^\top.
\end{equation}

From Equation~\eqref{eq:rel_SVD} we clearly see the correspondence of the
SVD  and the covariance of the data samples. The singular values of $X$ are the
square roots of the eigenvalues of the covariance matrix, and the singular
vectors of $X$ are the eigenvectors of the covariance matrix.

\subsubsection{Stochastic SVD}

In our application, the dataset is very large and in high dimensions, in which case
computing the eigenvalues of the data matrix is a key challenge. Recent
techniques enable the computation of large matrix decompositions in a  robust
manner without explicitly forming the entire data matrix in memory~\citep{bib:Halko2010,bib:Halko2011}.
The key main idea behind these methods is stochasticity.

For the construction of the SVD using stochasticity, a method we refer in this
manuscript as stochastic SVD is available~\citep{bib:Halko2011} which can be
broken in two stages. First, the construction of a subspace with reduced
dimension in which to represent the input data in an approximate manner; a
matrix is bult in which the columns form a reduced orthonormal basis for the data.
Second, the projection of the observations onto that subspace and the subsequent
computation of the SVD. Both stages are outlined in
Algorithm~\ref{alg:SVD_est}. 

\begin{algorithm}[h]
\scalefont{0.92}
\caption{-- Stochastic SVD\label{alg:SVD_est}}
{Given an ${m\times n}$ matrix $A$, and integers $l$ and $q$, compute an
approximate decomposition of $A\approx U\Sigma V^\top$, where $U$ and
$V$ are orthonormal and $\Sigma$ is a non-negative diagonal matrix.}
\begin{algorithmic}[1]  
    \State{Create an ${n\times l}$ matrix $\Omega$,  with independent entries
    and standard normal distribution.}
    \State{Form $Y=A\Omega$.}
    \State{Perform the $QR$ decomposition of $Y$,   $Y=QR$, with $Q$ an $m
    \times l$ orthonormal matrix.}
    \State{Form $B=Q^\top A$.}
    \State{Compute an SVD of the reduced matrix $B=\tilde{U}\Sigma V^\top$.}
    \State{Form the otrhonormal matrix $U=Q\tilde{U}$.}
\end{algorithmic}
\end{algorithm}

The stochastic SVD provides a robust means to perform large-scale matrix decompositions,
independently of the intrinisc structure of the data matrix. However, this
approach yields an approximate solution whose quality must be assessed in the
context of the original application. 

In this work, the purpose of the stochastic SVD is to obtain the principal
components of a given
dataset, \ie, to obtain the singular vectors to the left of an SVD of a data
matrix. We know that the obtained vectors form a basis for a subspace, so that
its precision can be assessed at each execution of the stochastic SVD. This
assessment was done using a distance between subspaces as described in the next
section.

\subsubsection{Distance between subspaces}\label{sec:subspaces}

Let    $F$ and $G$ be subspaces of $\mathbbm{R}^m$, with $p = \textsf{Dim} (F)
\geq \textsf{Dim} (G) \geq q \geq 1$. The princpal angles, $\theta_1, \dots,
\theta_q \in [0,\pi/2]$, between $F$ and $G$ are recursively defined as being real
the numbers $\theta_k$ such that 
\[
\cos(\theta_k) = \max_{u \in F,
v \in G} |u^\top v| = |u_k^\top v_k|,
\]
subject to 
\begin{equation*}
		u^\top u  = v^\top v = 1 , \qquad
		u^\top u_i = v^\top v_i = 0\ , \qquad i = 1,\dots, k-1 .
\end{equation*}
The vectors $\{ u_1, \dots, u_q\}$ e $\{ v_1, \dots , v_q\}$ are the so-called
principal vectors between the subspaces $F$ e $G$.

The greatest principal angle is related to the notion of distance between
subspaces of the same dimension~\citep{bib:Golub}. Thus, if $p = q$ then
$dist(F,G) = \sqrt{ 1 - \cos (\theta_p)^2} $ $ = \sin(\theta_p)$. In a practical
way, if the columns of $Q_F \in \mathbbm{R}^{m \times p}$ and those of $Q_G \in
\mathbbm{R}^{m \times p}$ define orthonormal bases for $F$ and $G$,
respectively, the cosine of the greatest principal angle is determined by
computing an SVD of the matrix $Q_F^\top Q_G$ and taking the smallest of its
singular values. It is important to note that
\begin{equation*}
0\leq dist(F,G) \leq 1.
\end{equation*}
The distance will be zero if $F = G$ and one if $F \cap G^{\perp} \neq \{0\}$,
with $G^{\perp}$ denoting the orthogonal complement of $G$, \ie, the space of all orthogonal vectors to $G$.

\subsection{Diffusion maps }

We have explored a more recent non-linear technique for dimensionality
reduction and manifold learning, the so-called diffusion maps, which re-organizes the data according to 
a reduced set of parameters related to the approximate intrinsic geometry of
the underlying phenomena~\citep{bib:Lafon,bib:Porte}. The reduced set of parametrs are computed from the
eigenvalues and eigenvectors of a diffusion operator on data. It is robust to
noise and outliers in data, and can be efficiently computed when its application is properly designed.

Consider that the set of
$n$ observations $y_1,\dots, y_n \in
\mathbbm{R}^p$ approximately sample one or more non-linear manifolds, each $y_i$ being a
vector of all pixels of a video frame, for instance. We describe the diffusion maps algorithm in four steps. The first step consists in building an $n \times n$ matrix $W$ of pairwise (local) similarities
between the observations. The similarities are defined by a kernel function $K :
Y\times Y \rightarrow \mathbbm{R} $, satisfying $K(y_i,y_j) =
K(y_j,y_i)$ and $K(y_i,y_j) \geq 0$. In the present work, we have used the heat
kernel given as
\begin{equation}
W_{ij} = K(y_i,y_j) =  e^{-\frac{|| y_i - y_j ||}{\epsilon}^2}.
\end{equation}
By choosing the parameter $\epsilon$ one can adjust the size of the neighborhood
with which to compute similarities, based on prior knowledge of the structure
and density of data~\citep{bib:Porte} and on sparsity considerations. In this work $\epsilon$ was taken as the
largest squared Euclidean distance between all datapoints, although a
significantly smaller
$\epsilon$ could have been chosen, leading to a sparser matrix $W$ thus increasing
the efficiency of the algorithm.

The second stage consists in constructing the diffusion matrix $P$, which is a stochastic
matrix, whose lines are normalized to 1.  The matrix $P$ is obtained by
\begin{equation}
P= D^{-1}W,
\end{equation}
where $D$ is a diagonal matrix whose entries are given by
\begin{equation}
D_{ii} = \sum_{j=1}^n W_{ij},  \quad \text{for } i = 1,\dots, n.
\end{equation}

Each entry of $P$ provides the
pairwise connectivity of the data in an underlying similarity graph. The graph
can be seen as a Markov
chain on the data points whose transition matrix is $P$;
each entry $P_{ij}$ represents the probability of
transitioning from data point $i$ to $j$ in one diffusion step. In other words,
this is the
probability of clustering together data points $i$ and $j$ in one step. When taking
powers of the transition matrix $P$, one increases the number of steps taken to
cluster nodes $i$ and $j$ to form a manifold.

The third step consists in computing the spectral decomposition of the
transition matrix $P$, thus obtaining the eigenvalues $\lambda_i$ and
corresponding eigenvectors $\psi_i$. Since the matrix $P$ is stochastic, its
greatest eigenvalue in absolute value, $\lambda_{0}$, equals $1$. When the
transition matrix is positive definite, $P$ has a sequence of positive
eigenvalues sorted in decreasing order
\begin{equation}
1 = \lambda_0 \geq \dots \geq \lambda_{n-1} > 0.
\end{equation}

The last step of diffusion maps is to actually perform the dimensionality
reduction. This is done by discarding eigenvalues of smaller indices.
In selecting the largest $q$ eigenvalues, we obtain a new feature vector
$\widetilde{y_j} \in \mathbbm{R}^q$, using the diffusion map given by
\begin{equation}
\widetilde{y_j} = 
\begin{bmatrix}
\lambda_1^t \psi_{j1}\\
\vdots	\\
\lambda_q^t \psi_{jq}\\
\end{bmatrix}\ ,
\end{equation}
with, \eg, $\psi_{j1}$ being the $j-$th component of eigenvector $\psi_1$ 
and $t > 0$ being a parameter corresponding to the power of the matrix $P$,
whose effect is to cluster through the Markov chain in $t$ steps.

\section{Image-Based Stochastic Modeling}

Our main goal is to devise a tractable stochastic model whose samples enable simulating a
video of the observed behavior of ocean surface patterns. The same approach
should be useful for image-based modeling and tracking of other continuous-time
deformations.  In addition to provide a framework for machine learning and inference, stochastic
modeling is necessary to account for aspects of the phenomenon that can be difficult or
impossible to model explicitly in an efficient manner.

For the automatic construction of the model from video, towards an application of recognizing ocean
patterns, we propose the use of Bayesian methods of probabilistic inference. The
use of these methods require a ``training'' or ``learning'' stage from enough input data and
then strategies for automatically fitting of the stochastic model 
to newly observed data, which we call the ``testing'' or ``tracking'' stage~\citep{bib:Mumford}.

The stochastic model will be built from the data-driven model given by either
diffusion maps or PCA.  Thus, consider that patterns of a signal are to be
modeled, where $S(t) \in \mathbbm{R}^q$ is the video frame at instant $t$ in
the diffusion map or PCA model space.  In this work, we specifically reduce the
frames to 3 dimensions, thus $q \in 3$, \ie, the motion 
patterns in the video are to be modulated by only 3 parameters.

Using the Bayesian inference paradigm, one seeks to infer the state of the
random variable $S(t)$ on a new time point given observable data $I(t)$ at other time points.
In this work some examples of observable variables are: 

\begin{itemize}
\setstretch{0.5} 
\item Number of sample points.
\item The local curvature between sample points.
\item The variation of sample time.
\end{itemize}

In order to build the stochastic model, it is necessary to define a probability
function $P(S(t),$ $ I(t))$. 
The inference of $S(t)$ is carried out using the \emph{a posteriori} probability
$P(S(t)|I(t))$ through Bayes' rule: 
\begin{equation}
P( S(t)|I(t) )  = \dfrac{P(S(t))P(I(t)|S(t))}{P(I(t))}, \text{ with } P(I(t)) = \sum_{S(t)} P(S(t))P(I(t)|S(t))
\end{equation}

The above general approach can lead to three problems~\citep{bib:Mumford}:
\begin{itemize}
\setstretch{0.5} 
\item The construction of the probability model, $P$;
\item To find an algorithm to maximize the \emph{a posteriori} probability;
\item To optimize the parameters $I$ of the model as to optimally fit the data.
\end{itemize}

In order to validate the stochastic model, one must take samples to produce a
stochastic simulation, and test if these reproduce the observed real-world flows
of the sea surface.  We have been actively working on adequate specific techniques for the above three
problems for the overall approach proposed here.

\section{Results and Discussion}

The experiments were preformed in the Scilab free software language and environment,
together with the image processing toolbox SIP.

\begin{figure}[!httb]
\centering
\includegraphics[scale=0.3]{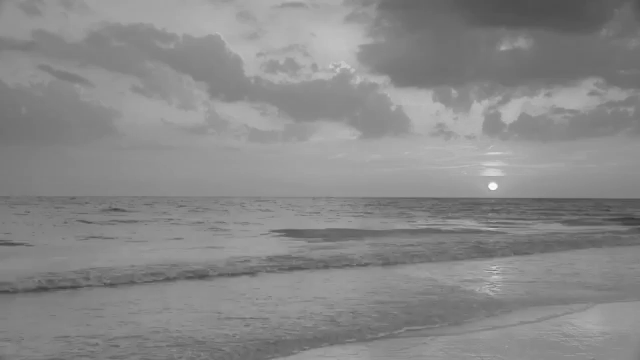}\\
\caption{Sample image of the video used in the experiments. The larger goal
is to obtain a model of the 3D structure and motion of the ocean surface. We
begin to tackle this by the construction of an efficient stochastic model of the
image patterns in the video.}
\label{fig:ExemploVideo}
\end{figure}

The data consisted of a video of a beach front
with a resolution of $360 \times 640$ and $21 \min 06 s$ duration,
illustrated in Figure~\ref{fig:ExemploVideo}.\footnote{\tiny{\it HD Florida
Beaches Sunset, Powerfloe Network},
\url{http://www.youtube.com/watch?v=0GBpGRXaruE}, 02/10/2012.  Used with permission.} 
This video was chosen since it provides a recording without camera movement, as
it simplifies the modeling of the problem without requiring a very large number of
frames. 

We extracted four $15s$ video clips to be analyzed from the original video,
forming the ``training set'' used to learn our model.
For each clip all frames of the video were extracted and converted to grayscale, resulting in a set of
$450$ images having $360 \times 640 =  230\,400$ pixels or dimensions. These
clips sample
the temporal evolution of the behaviour of ocean surface patterns. The
extracted frames of each clip go through dimensionality reduction using both
stochastic PCA and
diffusion maps in order to perform a comparative study of the 
power of each approach to represent the underlying patterns.

\paragraph{Reliability of the Stochastic PCA}
As we have used PCA with a stochastic SVD algorithm, the obtained subspace at each realization of
the method has a certain variance. In order to assess this, we 
computed the precision of the obtained subspaces for multiple runs of 
stochastic PCA, using the distance betweeen subspaces described in
Section~\ref{sec:subspaces}.

To the best of our knowledge, there is no information in the literature about
the number of samples in stochastic SVD needed to
obtain a set that generates a subspace that provides a good enough approximation
to the principal components.
For each set of images, the stochastic PCA was executed five times, followed by
the pairwise computation of subspace distances in order to assess the variance
of the result. Table~\ref{tab:dist_sub} shows the average distance between the
obtained subspaces, as well as the standard deviation for each video clip.

\begin{table}[!httb]
\centering
\caption{The average distance between the
subspaces obtained through multiple runs of stochastic PCA, together with the
standard deviation, for four different clips of the same original video.}
\label{tab:dist_sub}
\vspace{12pt}
	\begin{tabular}{|c||c c|}\hline
													  & \textbf{Average distance} & \textbf{Standard
                            deviation} \\ \hline\hline 
		\rule[-1ex]{0pt}{2.5ex} \textbf{Group 1} & 0.1099 & 0.003 \\  
		\rule[-1ex]{0pt}{2.5ex} \textbf{Group 2} & 0.1105 & 0.039 \\  
		\rule[-1ex]{0pt}{2.5ex} \textbf{Group 3} & 0.1064 & 0.056 \\  
		\rule[-1ex]{0pt}{2.5ex} \textbf{Group 4} & 0.1092 & 0.005  \\\hline
	\end{tabular} 
\end{table}

Despite the average distances between the subspaces at each execution of
stochastic PCA being close to $0.1$, which is around $10\%$ of the maximum
distances between two observed subspaces, this result is acceptable for the
applications up to this point. With the low standard deviation values, we have
confirmed that the proposed algorithm gives consistent results in practice.

\paragraph{Diffusion Maps Vs.\ Stochastic PCA}

\begin{figure}[!httb]
\centering
\subfigure[Stochastic PCA\label{fig:grupo1pca} ]{ \includegraphics[scale=0.26]{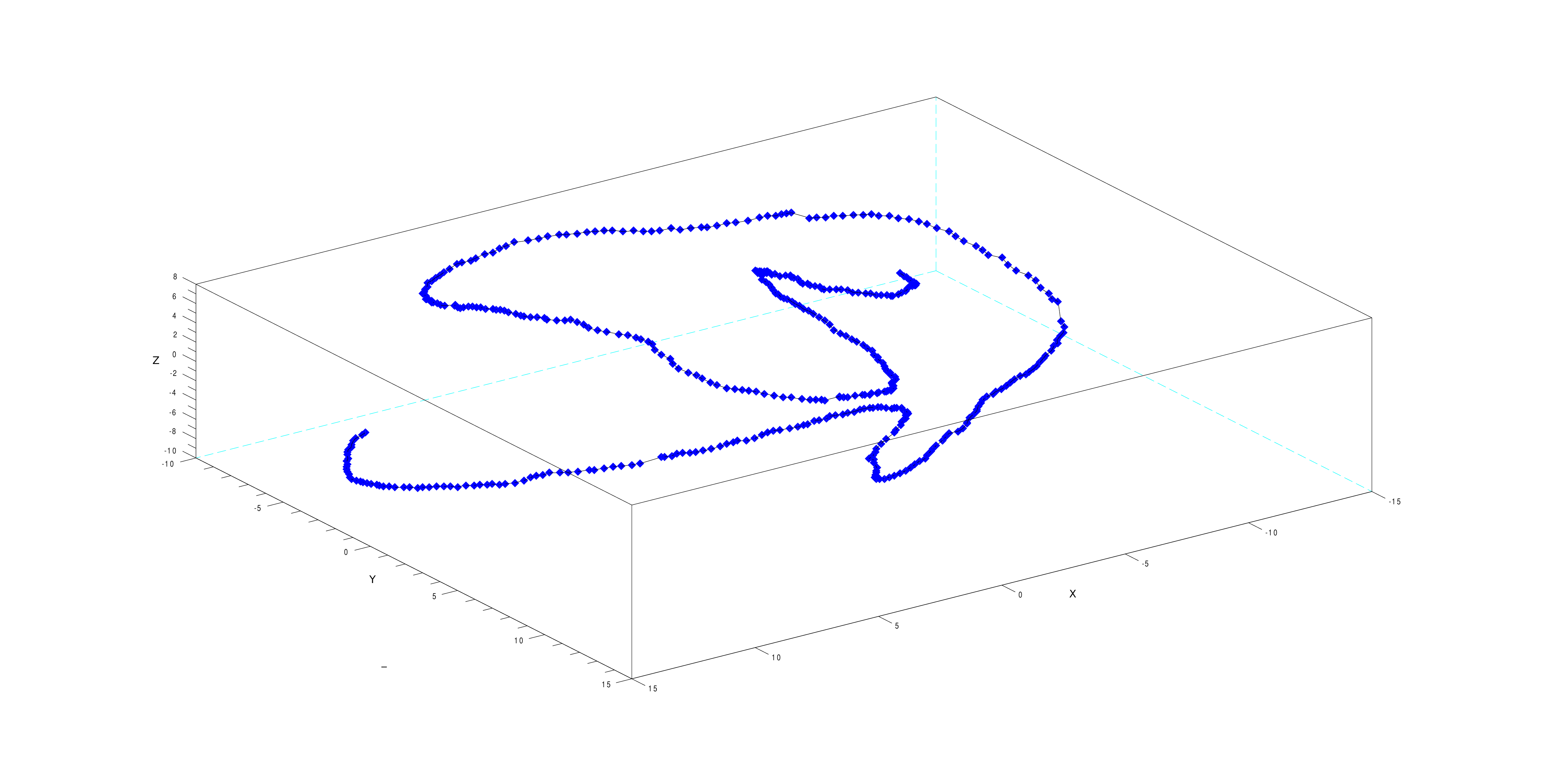}} 
\subfigure[Diffusion maps\label{fig:grupo1dif} ]{ \includegraphics[scale=0.26]{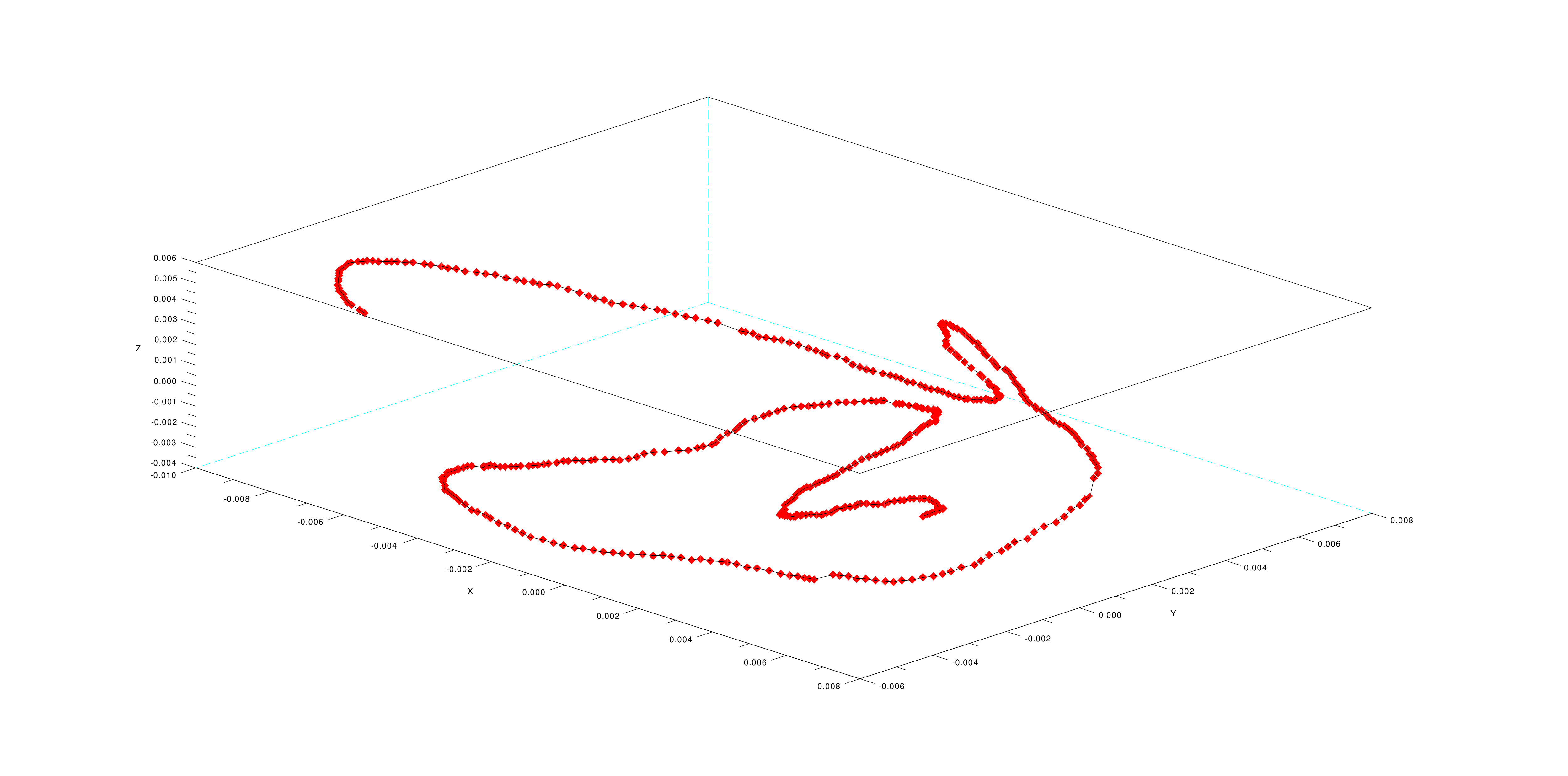}}
\caption{ Projection of each frame of the first videoclip ($15\min$) in
$\mathbbm{R}^3$ using~\subref{fig:grupo1pca}  one instance of stochastic PCA
and~\subref{fig:grupo1dif} diffusion map at scale $t=2$.}
\label{fig:exemplo:reducao}
\end{figure}

The results of applying dimensionality reduction using stochastic PCA and
diffusion maps for the first videoclip are shown in Figure~\ref{fig:exemplo:reducao}. 

\begin{figure}[!httb]
\centering
\includegraphics[scale=0.3]{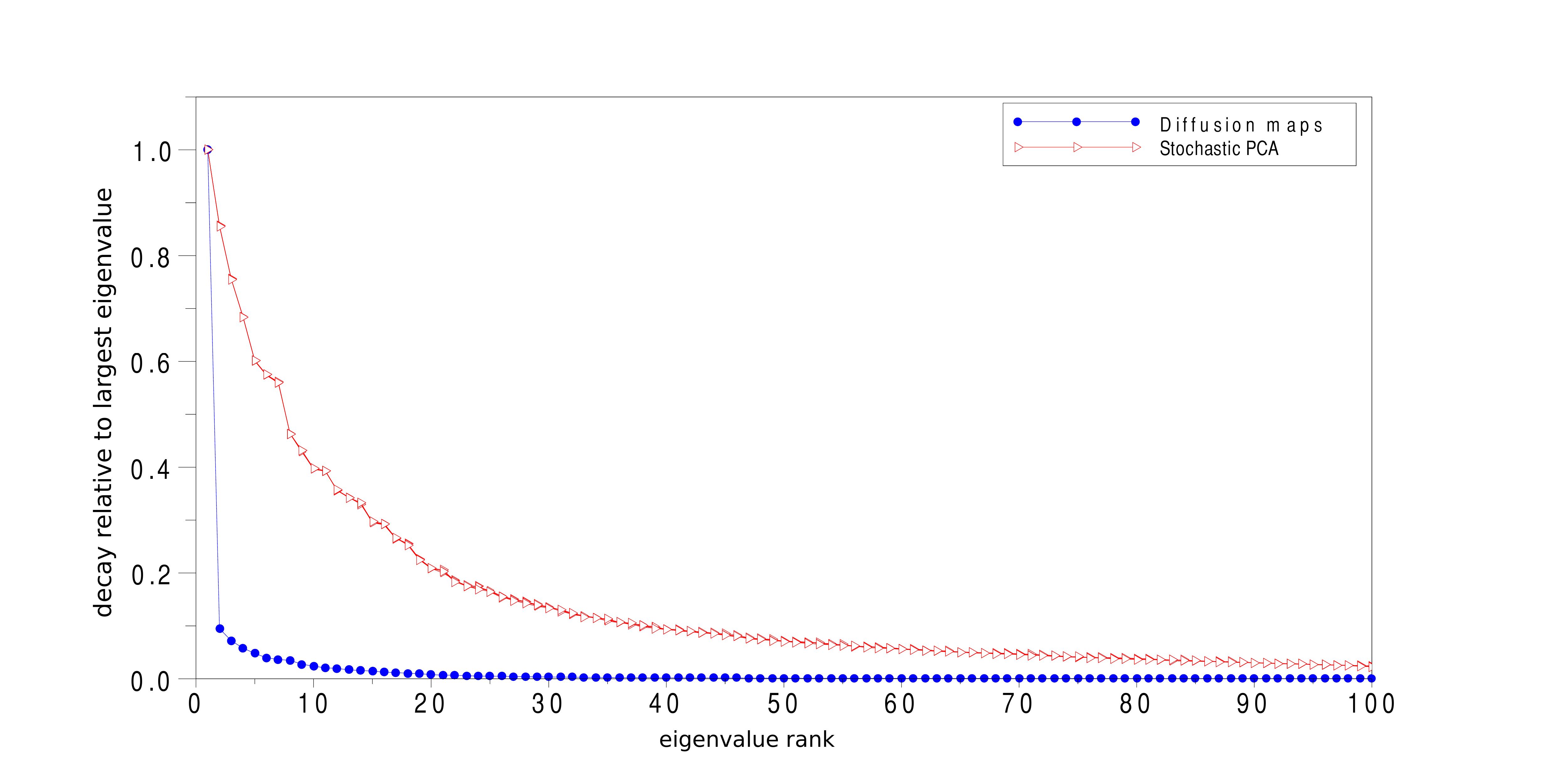}
\caption{The decay of the first $100$ eigenvalues used in the
diffusion map compared to that of the eigenvalues used in the stochastic PCA, relative
to the largest eigenvalues, for the first videoclip.}
\label{fig:queda:autovalor}
\end{figure}

From Figure~\ref{fig:exemplo:reducao} it is not possible to be conclusive about
which technique has better potential as part of the proposed modeling pipeline.
To further investigage this, we compare the decay of the eigenvalues used in the
diffusion map with that of the eigenvalues used in the stochastic PCA, relative
to the largest eigenvalues. The diffusion map eigenvalues present a much shaper
decay than the eigenvalues used in the stochastic PCA,
Figure~\ref{fig:queda:autovalor}. This indicates that diffusion maps can
represent the observed flow with fewer parameters.

\section{Conclusion and Future work}

The image-based fluid modeling approach proposed in this work is part of
a larger effort with applications to the online monitoring of ocean conditions
for aiding offshore processes of oil drilling, among others.

With dimensionality reduction it is possible to represent the original data with
a data-driven model with a small number of parameters. Using diffusion maps,
these parameters reveal intrinsic structure of the original data, keeping the
most relevant observable characteristics, while our experiments show that PCA
was not as efficient in this sense. Building on such reduced model we proposed the main ideas of
a stochastic model to account for remaining factors. 

Work is underway to produce techniques for detailed machine
learning and inference based on this model. The final complete model should be
able to capture the patterns of fluid surface motion in a way that, when
statistically sampling from this model, we should obtain a sequence of images that can
reproduce the observed fluid behavior through time. In the future this can then
be used to help infer the state of the ocean surface in a new video
sequence. We have also been considering the use of video streams from multiple
views to improve quality and robustness. 


\end{document}